# Discovery of Bias and Strategic Behavior in Crowdsourced Performance Assessment*

Yifei Huang
Outreach.io
Seattle, USA
yifei.huang@outreach.io

Matt Shum
California Institute of Technology
Pasadena, USA
mshum@caltech.edu

Xi Wu
Central University of Finance and Economics
Beijing, China
wuxi@cufe.edu.cn

Jason Zezhong Xiao
Cardiff University
Cardiff, UK
xiao@cardiff.ac.uk

## ABSTRACT

With the industry trend of shifting from a traditional hierarchical approach to flatter management structure, crowdsourced performance assessment gained mainstream popularity. One fundamental challenge of crowdsourced performance assessment is the risks that personal interest can introduce distortions of facts, especially when the system is used to determine merit pay or promotion. In this paper, we developed a method to identify bias and strategic behavior in crowdsourced performance assessment, using a rich dataset collected from a professional service firm in China. We find a pattern of "discriminatory generosity" on the part of peer evaluation, where raters downgrade their peer coworkers who have passed objective promotion requirements, while overrating their peer coworkers who have not yet passed. This introduces two types of biases: the first aimed against more competent competitors, and the other favoring less eligible peers which can serve as a mask of the first bias. This paper also aims to bring angles of fairness-aware data mining to talent and management computing. Historical decision records, such as performance ratings, often contain subjective judgment which is prone to bias and strategic behavior. For practitioners of predictive talent analytics, it is important to investigate potential bias and strategic behavior underlying historical decision records.

## 1 INTRODUCTION

Performance assessment is a fundamental aspect of human resources management (HRM). Accurate and informative performance assessment is highly valuable, as it serves a wide range of core HRM tasks: implementing incentive compensations such as merit pay, making critical personnel decisions such as promotions, and designing plans for workplace learning and career development [1, 14, 19]. In the realm of predictive talent analytics, outputs from performance assessment (e.g., performance ratings) are commonly used as input variables to predict career progression, turnover behaviors, job satisfaction, and other key outcomes [11, 17, 23, 27].

Assessing performance is never purely objective. Some subjective judgement is almost always involved, especially in evaluating the performance of creative workers. Traditionally, performance assessment uses a "top-down" system in which supervisors assess their subordinates [16]. However, since information of a specific employee's performance is dispersed among his/her supervisors, peers, subordinates, and even business partners, it is appealing to "crowdsource" performance assessment by asking all the relevant people to participate in performance assessment.

This is the basic idea behind crowdsourced performance reviews, 360-degree feedback, or multi-source assessment: employees' compensation and promotion decisions are partially based on performance assessment by themselves and their coworkers [5, 6]. With the industry trend of shifting from a traditional hierarchical approach to flatter organizational structure, crowdsourced performance assessment becomes more and more popular. It was reported that between one-third and one-half of U.S. companies use some form of crowdsourced performance assessment [22]. This is consistent with a separate estimation that over one-third of US companies and more than 90% of Fortune 500 firms use some form of 360-degree feedback [5, 10]. As an example, in Google, software engineers evaluate the performance of themselves and their co-workers every six months, and these assessments play an important role in merit pay and promotion decisions.[1]

While crowdsourcing performance assessment can have many advantages over traditional top-down approach, it also brings about new challenges, especially with regard to bias and strategic behavior. For instance, as noted by Jack Welch, former CEO of General Electric, "Like anything driven by peer input, the system is capable of being 'gamed' over the long haul" [25]. When the system is used to determine merit pay or promotion, raters likely face a conflict of interest problem in evaluating their colleagues, who are also potential competitors for promotions. Either wittingly or unwittingly, personal interest can introduce distortions of facts.

To this end, in this paper, we collected comprehensive HR archival data consisting of employees attributes, crowdsourced performance

---

1 Google's practice has been documented in Quora, a question-and-answer website: https://www.quora.com/How-are-performance-reviews-done-at-Google-What-are-they-used-for.

Discovery of Bias and Strategic Behavior in Crowdsourced Performance AssessmentTMC 2019, Alaska, USAassessment, and promotion outcomes, from a mid-sized Chinese professional service firm. We developed a method to discover and measure bias and strategic behavior. To our knowledge, this is one of the first public studies of crowdsourced performance assessment utilizing field data from a company.

The key of our approach is to leverage *objective measure of qualification* as a benchmark, to investigate how does the relationship between rater (i.e., the one who gives rating) and ratee (i.e., the one who receives rating) affect the rater's recognition of the ratee's qualification in assessment. We find that employees discriminate against "peers" (i.e., employees who are within the same hierarchical rank, and hence close competitors for promotions). Specifically, employees tend to denigrate qualified peers who have already passed objective requirements for promotion, while giving generous ratings to peers who have not yet passed. Employees recognize positive "qualification premium" to their nonpeer coworkers (i.e., giving higher rating to those qualified nonpeers than unqualified nonpeers), while recognizing negative "positive premium" to their peers. In addition, employees gave themselves favorable ratings so that on average an employee would improve her percentile rank by about 6.3% if the assessment result was dictated by her own ratings (to self and to coworkers in the team).

This paper also aims to bring angles of *fairness-aware data mining* [12, 18] to talent and management computing [24, 26, 28], where we are working on data produced by people about people for making high-stakes decisions directly impacting people. Historical decision records, such as performance ratings or promotion decisions, often contain subjective judgement which is prone to bias and strategic behavior. For practitioners of predictive talent analytics, it is important to investigate potential bias and strategic behavior underlying key variables derived from historical decision records.

## 2 DATA DESCRIPTION

The data used in our study were retrieved from a Chinese professional service firm's HR archive and performance assessment archive, covering a five-year period from 2010 to 2014. In the assessment, within each of the 7 engagement departments of the office, every employee is asked to evaluate everyone else within the department as well as to conduct self-evaluation. The maintenance of anonymity of each participant's evaluations is instructed in the assessment process.

Each observation in our dataset is a rating record, specifying the year of rating, the rater, the ratee, performance rating (averaged over the 30 dimensions), and information about the rater and ratee (e.g., department affiliation, rank at the time of performance evaluation, age, gender, educational background). We have a total of 7,778 rater-ratee-year observations for the five years comprising 153 unique employees in 7 departments of the firm. A 0-to-10 numeric scale of performance rating is used, where 0 indicates the poorest performance and 10 the best.

According to the firm's promotion guidelines, there are two requirements that an employee needs to meet in order to be qualified for promotion. First, the relative ranking of her performance assessment rating must be among the top 50% in the group of employees at the same level in her department. Second, for each level, there are some objective qualifications for promotion, including attendance, academic qualifications, project experience, and tenure.

## 3 METHODS OF DETECTING STRATEGIC BEHAVIOR

Our main empirical model is the following:

$$R_{ijt} = \beta_0 + \beta_1 Peer_{ijt} + \beta_2 Qual_{jt} + \beta_3 Peer_{ijt} \times Qual_{jt} + FE_{RateeRank} + FE_{Ratee} + FE_{Dept} + FE_{Year} + \varepsilon_{ijt} \quad (1)$$

In the regression equation (1), $R_{ijt}$ denotes the rating that rater $i$ gives to ratee $j$ at year $t$. The rating scale ranges from 0 to 10 with 0 denoting the poorest performance and 10 denoting the highest level of performance. We define the variable $Peer_{ijt}$ equal to 1 if the rater $i$ and ratee $j$ are of the same rank in year $t$, and 0 otherwise. We define the variable $Qual_{jt}$ equal to 1 if the ratee $j$ has already passed minimum requirements for promotion (in terms of attendance, academic qualifications, project experience, and tenure) at year t, and 0 otherwise. We control for fixed-effects specific to each ratee, ratee's rank, ratee's department, and evaluation year.

*Qualification Premium* captures the idea that a ratee who passed objective promotion requirements (*Qual=1*) should get higher ratings compared to the counterfactual scenario where the ratee does not pass (*Qual=0*). We define qualification premium as a function of *Peer*: *ΔQual(Peer) = R(PEER, 1) – R(Peer, 0)*. We can derive qualification premium conditional on whether rater is a peer:

$$\Delta Qual(Peer) = \begin{cases} R(1,1) - R(1,0) = \beta_2 + \beta_3 & \text{if } Peer=1 \\ R(0,1) - R(0,0) = \beta_2 & \text{if } Peer=0 \end{cases} \quad (2)$$

*Peer Difference* captures the difference in ratings that a ratee received from a peer rater (*Peer=1*) and that from a non-peer rater (*Peer=0*). We define peer difference as a function of *Qual*: *ΔPeer(Qual) = R(1, Qual) – R(0, Qual)*. We can derive peer difference conditional on whether ratee is qualified:

$$\Delta Peer(Qual) = \begin{cases} R(1,1) - R(0,1) = \beta_1 + \beta_3 & \text{if } Qual=1 \\ R(1,0) - R(0,0) = \beta_1 & \text{if } Qual=0 \end{cases} \quad (3)$$

How do we identify and measure strategic manipulation? Without manipulation, we expect peer raters and non-peer raters to behave similarly in recognizing qualification premiums. In other words, a rater's measured qualification premium should not depend on whether the ratee is a peer or not. That is, $\Delta Qual(1) = \Delta Qual(0)$, which implies that $\beta_3 = 0$. We can also consider this from the perspective of peer difference. Without manipulation, we expect the peer difference to be independent of whether the peer ratee has passed requirement or not. That is, $\Delta Peer(1) = \Delta Peer(0)$, which also gives to $\beta_3 = 0$. To sum up, the interaction terms in the regression model are important for quantifying strategic manipulation, and we expect $\beta_3 = 0$, in the absence of strategic manipulation.

## 4 RESULTS OF DETECTING STRATEGIC BEHAVIOR

### 4.1 Preliminary Evidence of Strategic Behavior

We start by providing some simple evidence showing that employees are indeed exhibiting self-interested manipulation in their rating behavior. We ask a simple question: how much do an employee's own ratings (of herself and of others) lead to a better performance ranking in the department than what she actually achieves in the assessment? In other words, how would an employee's assessment result be improved if the result was wholly dictated by her own ratings? To answer this question, we define a measure $\Delta PR_{self} = PR_{self} - PR_{actual}$, where $PR_{self}$ is the employee's percentile rank[2] according to her own rating and $PR_{actual}$ is her percentile rank in the actual assessment result. Since higher percentile rank corresponds to better relative ranking, a positive $\Delta PR_{self}$ implies that the employee's relative ranking according to her own ratings is better than what she actually achieves in the assessment. Table 2 presents the summary statistics of these three variables.

TABLE 1 ANALYSIS OF SELF-ASSESSMENT

| Variables | # Obs. | Mean | SD | Min | Q1 | Median | Q3 | Max |
|---|---|---|---|---|---|---|---|---|
| $\Delta PR_{self}$ | 432 | 0.065 | 0.171 | -0.625 | 0.000 | 0.037 | 0.140 | 0.847 |
| $PR_{self}$ | 432 | 0.573 | 0.300 | 0.000 | 0.338 | 0.611 | 0.833 | 1.000 |
| $PR_{actual}$ | 432 | 0.508 | 0.311 | 0.000 | 0.243 | 0.500 | 0.778 | 1.000 |

Notes: These summary statistics are computed over common observations.

The results suggest that an employee's percentile rank is substantially higher according to her own ratings, compared with the actual assessment result. **On average an employee would improve her percentile rank by about 6.5% if the assessment result was dictated by her own ratings**. Raters systematically rank themselves among the top half of employees in the department, thus placing themselves above the bar (which is at 50%) to satisfy the promotion requirement.

### 4.2 Ratings from Head, Peers, Nonpeers and Self

In the assessment process, each employee receives evaluations from his/her department head, peers, nonpeers and also conducts self-assessment. Do the average ratings of different components agree with each other? Table 1 presents the correlation matrix analysis.

TABLE 2 CORRELATION MATRIX: AVERAGE RATINGS FROM DIFFERENT COMPONENTS

| Component | All | Head | Peers | Nonpeers | Self |
|---|---|---|---|---|---|
| *All* | 1.000 | | | | |
| *Head* | 0.834*** | 1 | | | |
| *Peers* | 0.619*** | 0.520*** | 1 | | |
| *Nonpeers* | 0.982*** | 0.827*** | 0.490*** | 1 | |
| *Self* | 0.448*** | 0.439*** | 0.448*** | 0.409*** | 1 |

Notes: *, **, *** are significant at 10%, 5%, and 1%, respectively.

***Department head's ratings are less correlated to ratings from peers than to ratings from nonpeers*** (0.520 vs. 0.827). Interpreting department heads' ratings as a nonstrategic benchmark, this is consistent with our basic notion that peers are more likely to manipulate their ratings strategically than nonpeers. In addition, department heads' ratings and the overall average ratings are highly correlated (0.834).

Lastly, average ratings from nonpeers and the overall average ratings have a correlation coefficient as high as 0.982. This suggests that the peer evaluation part of the assessment only leads to a very limited degree of discrepancy between average nonpeer ratings and the overall ratings in our study.[3]

### 4.3 Regression Analysis: Ratee Qualification, Peer Status, and Strategic Rating Behavior

Table 3 presents the estimation results of the OLS regression. Overall, the empirical results suggest that raters do reward non-peer rates for passing the objective promotion requirements (i.e., $\beta_2 > 0$). However, the regression coefficient of the interaction term between *Peer* and *Qual* (i.e., $\beta_3$) is significantly negative, suggesting that **when the rater is a peer, she actually "punishes" the ratee who has passed promotion requirements** (i.e., the total effect is 0.165 - 0.493 < 0, $p < 0.01$). In other words, *ceteris paribus*, if the ratee passed promotion requirements, compared with the case of not passing, he/she would secure higher ratings from non-peer raters, and get lower ratings from peer raters. In addition, the coefficient on *Peer* (i.e., $\beta_1$) is significantly positive, **suggesting that raters give higher ratings to less qualified peers** (i.e., $\Delta Peer(0) > 0$).

TABLE 3 REGRESSION ANALYSIS: RATEE QUALIFICATION, PEER STATUS, AND STRATEGIC RATING BEHAVIOR

| Dep. Var.: $RATING_{ijt}$ | | OLS |
|---|---|---|
| $Peer_{ijt}$ | $\beta_1$ | 0.436*** |
| | | (0.061) |
| $Qual_{jt}$ | $\beta_2$ | 0.165*** |
| | | (0.042) |

---

[2] If there are n people and an employee is ranked as the k-th highest, then her percentile rank is (k-1)/(n-1). She gets a percentile rank of 1 if she obtains the highest rating, while a percentile rank of 0 corresponds to the poorest rating in the department.

[3] These results remain robust if we use within-department percentiles of average ratings instead of the raw value of average ratings.

| | | |
|---|---|---|
| $Peer_{ijt} \times Qual_{jt}$ | $\beta_3$ | -0.493*** |
| | | (0.079) |
| Ratee rank fixed-effects | | Yes |
| Ratee fixed-effects | | Yes |
| Year fixed-effects | | Yes |
| Department fixed-effects | | Yes |
| $N$ | | 7,346 |
| $Adj.\ R2$ | | 0.393 |

Notes: Robust standard errors clustered by ratee and year are reported in parentheses. *, **, *** are significant at 10%, 5%, and 1%, respectively (two-tailed).

In Figure 1 we illustrate the Qualification Premium and Peer Differences, using estimated coefficients from Table 3. We can clearly see the interaction effects between ratee qualification and peer status. Qualification premium is negative when the rater is a peer, while it is positive when the rater is a nonpeer. Peer difference is positive when the ratee failed qualification, while it becomes slightly negative when the ratee is qualified.

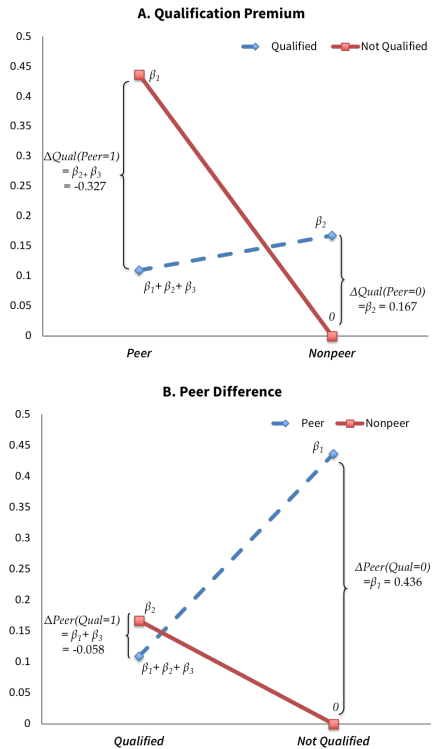

**Figure 1 Ratee Qualification Premium and Peer Difference**

These results indicate a **"discriminatory generosity"** on the part of peer raters, leading them to denigrate the relative ranking of peer ratees who have already passed promotion requirements, while overrating peer ratees who have not yet passed promotion criteria. This is consistent with our notion of strategic reporting, which simultaneously introduces *two types of biases: the first aimed against more competent competitors, and the other favoring less eligible peers (possibly as a mask of the first bias)*.

## 5 FURTHER ANALYSIS: EVALUATING THE IMPACT OF STRATEGIC BEHAVIOR

In this session, we further explore the relationship between performance assessment and promotion decisions. Specifically, we are interested in learning to which extent the strategic behavior in peer evaluation affect promotion outcomes. This counterfactual reasoning is very challenging; we consider our attempts here as an exploration to generate some insights rather than fully solving the problem.

### 5.1 Ratings and Promotion Decisions

The first step in answering this question is to estimate the effects of the performance ratings on employees' promotion probabilities. To do this, we collected annual promotion outcomes from the firm's personnel archive. Then we estimate empirical specifications to determine how much good performance ratings and passing objective promotion requirements affected an employee's chances of being promoted within the firm.

Table 4 presents estimation results of our logistic regression models using $PROMOTION_{jt}$ as the dependent variable (coded 1 if a ratee gets promoted after the annual performance assessment, and 0 otherwise). In these regressions, we use as a regressor $PR_{jt}$, the within-department percentile of an employee's average rating, rather than the raw rating. We do so for two reasons. First, the firm uses relative rankings to specify the minimum requirement for being considered for promotion. Second, the rating percentile provides a more comparable measure across years, since it is invariant to fluctuations of rating leniency over the five years of our sample. In addition, we include in the model $LICENSE_{jt}$ (coded 1 if the ratee $j$ has obtained the license in year $t$, and 0 otherwise) to control for differences in professional qualifications across ratees. Finally, we include fixed effects for year, department, and (pre-evaluation) rank.

TABLE 4 LOGISTIC REGRESSION ANALYSIS OF PROMOTION DECISIONS

| Dep. Var.: $PROMITION_{jt}$ | Logit | Logit |
|---|---|---|
| $PR_{jt}$ | 6.721*** | 5.472*** |
| | (1.235) | (1.088) |
| $Qual_{jt}$ | 1.546*** | 0.490 |
| | (0.555) | (0.324) |
| $PR_{jt} \times Qual_{jt}$ | -2.572** | |
| | (1.070) | |
| $License_{jt}$ | 1.373*** | 1.282*** |
| | (0.393) | (0.391) |
| Constant | 3.408*** | -2.850*** |
| | (0.826) | (0.784) |
| Rank fixed-effects | Yes | Yes |
| Year fixed-effects | Yes | Yes |
| Department fixed-effects | Yes | Yes |
| $N$ | 426 | 426 |
| $Pseudo\ R^2$ | 0.380 | 0.370 |

Notes: Standard errors are reported in parentheses. *, **, *** are significant at 10%, 5%, and 1%, respectively (two-tailed).

In Table 4, Column 1 presents the specification which includes the percentile of rating ($PR_{jt}$), ratee qualification ($Qual_{jt}$) and the interaction term between them. The coefficients on $PR_{jt}$ and $Qual_{jt}$ are both positive and significant at 1%. The interaction term is negative and significant at 5%. Column 2 presents the specification without the interaction term. While the coefficient on $Qual$ remains positive, the statistical significance weakened (p-value = 0.13).

These results suggest, first, that passing the objective qualification contributes to promotion. Second, the negative coefficient on the interaction term indicates that the marginal importance of performance rating decreases as the employee passes promotion requirements. In other words, a good rating is more important for those who have not yet passed promotion qualifications. These results suggest substitutability between performance rating and passing promotion requirements.

## 5.2 Counterfactual Simulation: Impact of Strategic Behavior on Promotion

Logically, we expect strategic behavior in peer evaluation to distort the aggregated assessment results in a direction that benefits those who have not yet passed promotion requirements, who manipulated ratings to improve their relative ranking among their peers. Whether and to what extent the strategic manipulation biases assessment results and promotion outcomes is a question of significant practical implications. Here we conduct explorations to generate some insights using counterfactual simulation.

There are four counterfactual scenarios to consider.

1. $CS_{head}$ Assessment results are determined only by the department head's ratings. This proxies for the rating that an employee would have received in a hypothetical top-down performance evaluation scenario.
2. $CS_{peer}$ Assessment results are determined only by the peer evaluation component.
3. $CS_{nonpeer}$ Assessment results are determined only by the nonpeer component. All peer evaluations in the original assessment are dropped. This helps us to see to what extent the peer evaluation part distorts overall promotion outcomes.
4. $CS_{self}$ Assessment results are determined only by self-assessments. An employee's percentile rank is determined by the relative ranking of her self-assessment compared with the ratings she gives to others.

In each scenario, we will use the corresponding counterfactual assessment results to simulate the probability of promotion (denoted as $P_{CS}$) for each employee, employing the fitted logistic model of promotion decision in the first column of Table 4. Then, we compare the counterfactual promotion probabilities with actual promotion probabilities (denoted as $P_{actual}$) which are the predicted promotion probabilities using the actual assessment results. We define $\Delta_{CS} = P_{CS} - P_{actual}$ as the increase of promotion probability in the counterfactual scenario compared with the actual case. For example, if an employee's $P_{actual}$ is 0.3 and $P_{head}$ is 0.4, then her $\Delta_{head}$ is 0.1. That is, her probability of promotion would increase by 0.1 if the promotion decision was solely determined by her department head's rating.

Table 5 presents the summary statistics of the counterfactual changes in promotion probabilities. The most notable result is that employees increased their promotion probabilities if the appraisal results were determined by their own ratings. In this scenario ($CS_{self}$), 42.79% of employees increased their promotion probabilities while 31.84% of employees had decreased promotion probabilities. On average, an employee increased her promotion probability by 3.98 percentage points under $CS_{self}$, which represents a relative increase of 7.16% (the average probability of actual promotion is 55.59% in this sample). This result characterizes the extent to which one's self-assessment promotes self-interest. In fact, this result is driven by that self-assessment tends to inflate one's own percentile rank as showed in Table 1.

TABLE 5 SUMMARY STATISTICS: SIMULATED CHANGES IN PROMOTION PROBABILITY

| Variables | # Obs. | Mean | Median | SD | % negative | % zero | % positive |
|---|---|---|---|---|---|---|---|
| $\Delta_{head}$ | 426 | 0.0125 | 0 | 0.1060 | 38.97 | 28.64 | 32.39 |
| $\Delta_{peer}$ | 368 | 0.0079 | 0 | 0.1566 | 37.50 | 24.18 | 38.32 |
| $\Delta_{nonpeer}$ | 426 | 0.0003 | 0 | 0.0677 | 27.46 | 47.18 | 25.35 |
| $\Delta_{self}$ | 402 | 0.0398 | 0 | 0.1398 | 31.84 | 25.37 | 42.79 |

Table 6 presents the correlation matrix of these four simulated changes in promotion probabilities together with ratee qualification. First, $\Delta_{peer}$ and $\Delta_{nonpeer}$ have a significantly negative correlation coefficient of −0.283. This is the only pair of $\Delta_{CS}$ variables that are negatively correlated, suggesting that these two counterfactual scenarios lead to different consequences upon promotion outcomes. This result is consistent with our expectation that peer evaluation and nonpeer evaluation reflect different motives in the rater. Second, $\Delta_{peer}$ and $Qual$ are negatively correlated (−0.103, and significant at 5%), which is consistent with our earlier finding that raters gave higher ratings to their unqualified peers while gave lower ratings to their qualified peers.

TABLE 6 CORRELATION MATRIX: RATEE QUALIFICATION AND SIMULATED CHANGES IN PROMOTION PROBABILITY

| Variables | $Qual$ | $\Delta_{head}$ | $\Delta_{peer}$ | $\Delta_{nonpeer}$ | $\Delta_{self}$ |
|---|---|---|---|---|---|
| $Qual$ | 1.000 | | | | |
| $\Delta_{head}$ | 0.017 | 1 | | | |
| $\Delta_{peer}$ | -0.103** | 0.126** | 1 | | |
| $\Delta_{nonpeer}$ | -0.061 | 0.178*** | -0.283*** | 1 | |
| $\Delta_{self}$ | 0.033 | 0.304*** | 0.218*** | 0.165*** | 1 |

## 5 RELATED WORK

In this session, we review some related work in economics, management science, and accounting on performance assessment.

This aims to facilitate interdisciplinary perspective which is particularly valuable for talent and management computing.

***Subjective performance evaluation and optimal incentive contract.*** In economics, there is a large body of theoretical literature on optimal incentive contracting with both objective measures and subjective evaluations [1, 19, 20]. Some recent studies also consider the role of crowdsourced performance evaluation [8, 9].

***Tournament, peer performance evaluation, and sabotage.*** In economics and management science, researchers conducted laboratory experiments to simulate workplace environment. In a setting where peer performance assessment is used to determine the allocation of tournament prize, participants exert less effort when sabotage becomes more likely, recognizing that their performance would not be fairly recognized by their peers [7]. Tournament structures (or relative performance schemes in general) have the potential to incentivize higher effort, it also induces higher sabotage, which can reverse the incentive effects [2, 13].

***Bias in traditional top-down performance assessment.*** Subjective performance evaluation has also been a long standing research topic in management accounting. However, the focus of the literature has been on the traditional top-down managerial appraisal of subordinates [3, 4, 15, 21]. [3] shows that both centrality bias (raters' tendency to compress performance ratings) and leniency bias (tendency to inflate employees' performance ratings) in performance evaluation are managers' defensive mechanisms to alleviate ramifications of truthful ratings. And both centrality bias and leniency bias are positively affected by information gathering costs and strong employee-manager relationships, but they do not necessarily damage employee performance.

## 6  CONCLUSION

In this paper, we investigated bias and strategic behavior in crowdsourced performance assessment, using a rich dataset collected from a professional service firm in China. We developed a method that analyzes how the relationship between rater and ratee affects the recognition of the ratee's objective measure of qualification. We find that a pattern of "discriminatory generosity" on the part of peer raters, who denigrate the relative ranking of peer ratees who have already passed promotion requirements, while overrating peer ratees who have not yet passed promotion criteria. We also explored the impact of the bias and strategic behavior using counterfactual simulation based on the relationship between performance assessment and promotion outcomes.

## ACKNOWLEDGMENTS

We thank Ming Hsu, Lawrence Jin, Clive Lennox, Jian Ni, Alejandro Robinson, Jean-Laurent Rosenthal, Thomas Ruchti, Robert Sherman and participants in presentations at Caltech and Zhejiang University for useful comments. Xi Wu thanks the managing partner and the head of the human resource department of the participating professional service firm for providing proprietary data and information on performance evaluation that make this study possible.